\documentclass[letterpaper, 10 pt, conference]{ieeeconf}

\IEEEoverridecommandlockouts
\usepackage{hyperref}
\hypersetup{colorlinks=true,
linkcolor=blue,
anchorcolor=blue,
citecolor=blue}

\usepackage{graphicx}
\usepackage{amsmath}
\usepackage{amssymb}
\usepackage{booktabs}
\usepackage{xcolor}
\usepackage{cite}
\usepackage{pifont}

\title{\LARGE \bf
Self-Supervised Mask-Aware Transformers for Fault-Tolerant FBG Force Sensing in Minimally Invasive Surgical Robotics
}

\author{Peibo Sun$^{1,\dagger}$, Shiyuan Dong$^{1,\dagger}$, Shucheng Ye$^{1}$, Jianrong Cai$^{1}$, Yushan Liu$^{2}$,\\ Hongen Liao$^{1}$, Tianqi Huang$^{1}$, Fang Chen$^{1,*}$%
\thanks{This work was supported by the National Key Research and Development Program of China (2025YFC2426300), the National Natural Science Foundation of China (Grant Nos. 82572314, 62477031, 62403307, and 62271246), the Science and Technology Commission of Shanghai Municipality (Nos. 24511104100, 25ZR1402225, and 24ZR1439800), and the Open Research Fund of the State Key Laboratory of Multimodal Artificial Intelligence Systems.}%
\thanks{$^{1}$P. Sun, S. Dong, S. Ye, J. Cai, H. Liao, T. Huang, and F. Chen are with Shanghai Jiao Tong University, Shanghai, China.}% 
\thanks{$^{2}$Y. Liu is with Shenzhen International Graduate School, Tsinghua University, Shenzhen, China.}%
\thanks{$^{\dagger}$Equal contribution. }%
\thanks{$^{*}$Corresponding author: Fang Chen ({\tt\small chen-fang@sjtu.edu.cn}).}%
}

\begin{document}
\maketitle
\thispagestyle{empty}
\pagestyle{empty}

\begin{abstract}
In minimally invasive surgical robotics, catheter-scale Fiber Bragg Grating (FBG) sensors are promising due to their ability to estimate multi-dimensional forces by multiplexing several optical channels. However, deploying these compact multi-channel sensors introduces two critical engineering challenges: inherent nonlinear cross-axis coupling during complex deformations, and intermittent channel dropouts caused by fiber fractures in constrained workspaces. These compounding issues severely degrade force estimation. Existing fault-tolerant approaches rely on combinatorial model banks, which scale exponentially with the channel count and demand prohibitively expensive per-pattern calibration. In this paper, we propose a unified, self-supervised mask-aware Transformer that explicitly models channel availability to enable graceful degradation under diverse and dynamic sensor failures. The encoder is pretrained via masked-channel reconstruction on unlabeled data streams and fine-tuned for force regression using a balanced clean-and-corrupted-view objective alongside a dynamic corruption curriculum. Furthermore, a parallel uncertainty head, trained via heteroscedastic Gaussian negative log-likelihood, predicts per-axis confidence in a single forward pass, circumventing the overhead of multi-pass ensembles. Evaluated on a catheter-scale 8-channel FBG dataset, our single unified model achieves a nominal Root Mean Square Error (RMSE) of 0.0066~N and degrades gracefully to 0.0126~N under severe 4-channel failures. This significantly outperforms a comprehensive model bank of 255 per-pattern neural networks (0.0154~N at 4-channel loss) while eliminating pattern-specific calibration. Finally, we demonstrate that the predicted uncertainty strongly correlates with physical ill-conditioning, establishing a reliable $\tau$--$\delta$ safety contract to mitigate risks in force-controlled surgical interventions.
\end{abstract}

\section{Introduction}
Reliable distal force sensing is essential for safe tissue interaction in minimally invasive surgery (MIS) \cite{gan2021development,patel2022haptic,lai2025fbg}. Fiber Bragg Grating (FBG) force sensors have emerged as strong candidates for MIS due to their inherent compactness, biocompatibility, and magnetic resonance compatibility \cite{wu2020fbg,dong2025high,lou2024fbg,dong2026fiber}. However, deploying catheter-scale FBG sensors introduces two critical engineering challenges: (i) strong nonlinear coupling among channels due to extremely tight manufacturing tolerances \cite{deng2020miniature,fu2025multifunctional}, and (ii) intermittent \emph{channel failures} caused by connector fatigue or partial optical fiber fractures during repeated bending in tortuous surgical pathways \cite{li2022three}. These failures are often correlated because multiple gratings are multiplexed on the same optical fiber and therefore share common physical vulnerabilities. Consequently, a robust force estimator must achieve high accuracy under nominal conditions while degrading gracefully when varying subsets of channels become unreadable.

\begin{figure*}[t]
  \centering
  \includegraphics[width=0.9\linewidth]{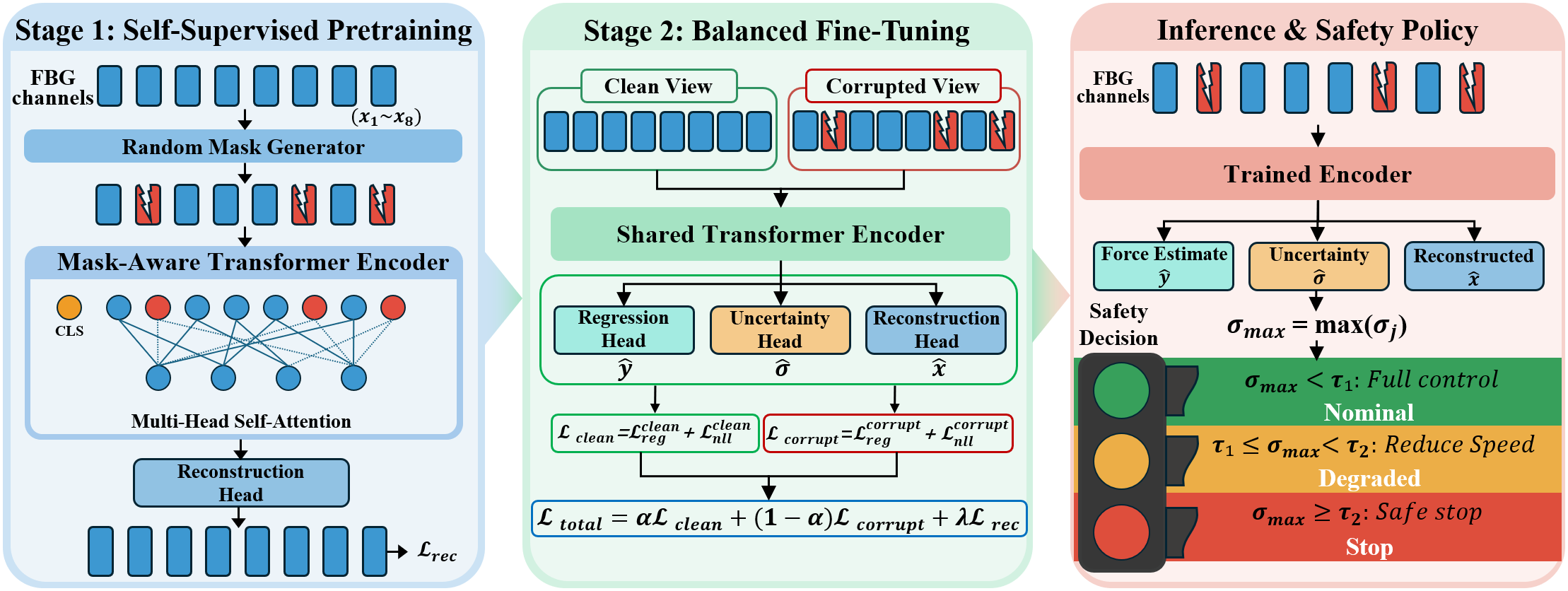}
  \caption{Overview of the proposed training pipeline: self-supervised masked reconstruction pretraining followed by balanced fine-tuning for force regression.}
  \label{fig:method_overview}
  \vspace{-7mm}
\end{figure*}

To address the highly nonlinear cross-axis coupling in compact structures, data-driven decoupling approaches have largely superseded classical linear or polynomial calibration \cite{gao2018fiber,hao2024development,dong2024high}. However, standard neural networks rigidly assume fully observed inputs and suffer catastrophic performance collapse under channel dropouts. To achieve fault tolerance, recent FBG methods \cite{li2023fault,li2024modular} heavily rely on a \emph{model-bank} strategy—calibrating and maintaining a dedicated model for every possible failure pattern. This paradigm scales exponentially with the number of channels $C$, requiring up to $2^C{-}1$ discrete models. Such an explosion in model count---amounting to 255 models for an 8-channel sensor---renders the approach computationally prohibitive in terms of onboard memory footprint for high-density arrays. Furthermore, switching between discrete models introduces non-trivial pattern-matching latency during real-time control loops, severely hindering high-frequency robotic interventions. Moreover, while risk-aware control is imperative in surgical robotics~\cite{kendall2017uncertainties}, existing uncertainty quantification techniques, such as Monte Carlo (MC) Dropout \cite{gal2016dropout} or Deep Ensembles \cite{lakshminarayanan2017simple}, require computationally expensive multiple forward passes and fail to explicitly model the physical degradation of the sensor structure.

Beyond FBG-specific model banks, broader learning-based robotic force estimators \cite{roshanfar2025learning} and traditional interacting multiple-model (IMM) fault-tolerant architectures \cite{kheirandish2023fault} similarly fail to scale with high sensor density. Recently, however, masked autoencoders (MAEs) have emerged as powerful paradigms for handling missing data in fragmented wearable \cite{xu2025lsm} and industrial sensor streams \cite{fan2024multiscale}. Inspired by these cross-disciplinary successes, we hypothesize that MAEs can replace rigid multi-model schemes.

To overcome these fundamental limitations, specifically the combinatorial explosion of calibration models and the inability to generalize to unseen failure patterns, we propose a \emph{single unified architecture} that adapts the masked modeling paradigm—successful in vision and language domains \cite{he2022mae,devlin2019bert}, and increasingly in incomplete multi-sensor time-series \cite{xiang2026learning}—to structured tactile sensor arrays \cite{zerveas2021transformer,cheng2024maskcae,dong2026fbg}, treating channel dropouts as masked inputs. By directly consuming an explicit observation mask, our network accommodates diverse and dynamic channel failures in a single inference pass, eliminating the exponential overhead and latency of per-pattern model switching. 

\textbf{Contributions:} 
\begin{itemize}
    \item To address the fault-tolerance challenge in multi-channel hardware, we propose a unified mask-aware Transformer that replaces the combinatorial $2^C{-}1$ model bank. By incorporating observation masks via multiplicative gating and attention key-padding, the architecture reliably handles dynamic and variable missing channels with zero-overhead pattern generalization.
    \item We design a two-stage training scheme: self-supervised masked-channel reconstruction on unlabeled data to capture physical intra-fiber correlations, followed by balanced fine-tuning using a dynamic corruption curriculum for robust force regression.
    \item We introduce a single-pass heteroscedastic uncertainty head to predict per-axis confidence. We empirically validate its reliability through condition-number analysis and deploy it to establish a rigorous $\tau$--$\delta$ safety contract, enabling active risk-mitigation in surgical settings.
\end{itemize}

\section{Method}

\textbf{Problem Formulation.} For the custom 8-channel FBG sensor developed in this study, let $\mathbf{x}\in\mathbb{R}^8$ denote the raw wavelength shifts and $\mathbf{y}\in\mathbb{R}^3$ the target three-axis tip force. To explicitly model the physical availability of each sensor channel, we define a binary observation mask $\mathbf{m}\in\{0,1\}^8$, where $m_i{=}1$ indicates an intact, readable channel, and $m_i{=}0$ denotes a failed channel. Our objective is to learn a unified, fault-tolerant mapping $\hat{\mathbf{y}} = f_\theta(\mathbf{x}, \mathbf{m})$ that maintains high nominal accuracy when $\mathbf{m}=\mathbf{1}$ and degrades gracefully for diverse and dynamic failure patterns.

Figure~\ref{fig:method_overview} summarizes the proposed unified architecture and the two-stage training pipeline.

\subsection{Input Representation and Failure Models}
\label{sec:input_failure}
Our tubular sensor architecture physically multiplexes eight FBG channels into four optical fibers, resulting in four distinct channel pairs: (1--2), (3--4), (5--6), and (7--8). Because gratings located on the same fiber share identical strain paths and structural vulnerabilities, a single physical fracture inherently invalidates its associated channel pair simultaneously. To ensure the learned representations are robust against these hardware realities, we introduce two complementary failure models during both training and evaluation:
\begin{itemize}
    \item Random $k$-channel dropout: Masks $k$ uniformly selected channels ($k=\lVert \mathbf{1}-\mathbf{m} \rVert_1$) to simulate independent, intermittent optoelectronic or connector faults.
    \item \looseness=-1 Fiber-level dropout: Masks predefined channel pairs simultaneously to model catastrophic physical fiber fractures.
\end{itemize}

\looseness=-1
At inference, missing channels ($m_i{=}0$) are initially imputed with the training-set mean. All channels are subsequently standardized (zero-mean, unit-variance) to yield normalized inputs $\bar{x}_i$. The binary mask $\mathbf{m}$ is injected into the network as an explicit condition, enabling the model to differentiate between valid measurements and imputed placeholders.

\subsection{Mask-Aware Transformer Encoder}
\looseness=-1
Instead of relying on rigid fully connected layers, we process the sensor array as a structured sequence. Each channel is represented by a token derived from the gated, standardized measurement $\tilde{x}_i = m_i \bar{x}_i$. A linear projection maps the scalar $\tilde{x}_i$ to a $d$-dimensional embedding, which is then summed with a learnable channel-specific positional embedding.

We apply multi-head self-attention \cite{vaswani2017attention} combined with a key-padding mask derived from $\mathbf{m}$. This guarantees that missing channels do not contribute as attention keys or values, preventing imputed noise from corrupting the representations of healthy channels. The encoder ultimately yields a global state for force regression and per-channel tokens for auxiliary reconstruction.

We exploit the known hardware topology by augmenting the channel-level token sequence with multi-scale fiber representations. Letting $\mathcal{G}_j$ denote the index set of channels multiplexed on fiber $j$, we compute the corresponding fiber token via masked mean pooling:
\begin{equation}
\mathbf{t}^{(f)}_j = \mathbf{W}_f \left(\frac{\sum_{i\in \mathcal{G}_j} m_i \bar{x}_i}{\sum_{i\in \mathcal{G}_j} m_i + \epsilon}\right) + \mathbf{e}_f(j),
\end{equation}
where $\mathbf{W}_f$ represents a linear projection matrix, $\mathbf{e}_f(j)$ provides the fiber-specific positional embedding, and $\epsilon$ is a small constant preventing division by zero. By integrating these multi-scale tokens, the self-attention mechanism fuses mechanical strain information across both individual gratings and holistic fiber structures.

\subsection{Self-Supervised Pretraining by Masked Reconstruction}
We employ self-supervised masked modeling \cite{he2022mae} to learn intrinsic cross-channel physical correlations without requiring expensive force labels. Given fully observed training inputs $\mathbf{x}$, we artificially corrupt the sequence by hiding a random subset of channels. Let $\tilde{\mathbf{m}}$ be the artificially corrupted mask and $\hat{\mathbf{x}}$ the network's reconstruction output. We optimize a masked Mean Squared Error (MSE) exclusively on the hidden channels:
\begin{equation}
\mathcal{L}_{\mathrm{rec}} = \frac{1}{\lVert \mathbf{m}{-}\tilde{\mathbf{m}} \rVert_1} \sum_{i=1}^8 (\hat{x}_i - \bar{x}_i)^2 \cdot (m_i - \tilde{m}_i).
\end{equation}
This forces the encoder to capture the underlying mechanical coupling (e.g., how bending in the $X$-axis affects opposite fibers), yielding a resilient weight initialization for the subsequent regression task.

\subsection{Balanced Fine-Tuning for Robust Force Regression}
\looseness=-1
During the supervised fine-tuning stage, we attach a regression head to the classification (CLS) token to predict the three-axis force $\mathbf{y}$ using a robust Huber loss ($\mathcal{L}_{\mathrm{reg}}$). To simultaneously maximize nominal precision and fault tolerance, we compute the loss on both a \emph{clean view} and a \emph{corrupted view} of the same batch:
\begin{equation}
\begin{aligned}
\mathcal{L}_{\mathrm{task}} = {} & \alpha\,\mathcal{L}_{\mathrm{reg}}\bigl(f_\theta(\mathbf{x},\mathbf{m}),\mathbf{y}\bigr) \\
& + (1{-}\alpha)\,\mathcal{L}_{\mathrm{reg}}\bigl(f_\theta(\mathbf{x},\tilde{\mathbf{m}}),\mathbf{y}\bigr) + \lambda\,\mathcal{L}_{\mathrm{rec}}.
\end{aligned}
\end{equation}
We use a balanced weighting ($\alpha{=}0.5$) and a small auxiliary reconstruction weight $\lambda$. To prevent underfitting during the early stages, the corrupted view $\tilde{\mathbf{m}}$ is generated using a \emph{dynamic corruption curriculum}: at epoch $e$, the number of masked channels is sampled as $k\sim\mathcal{U}\{0,\dots,k_{\max}(e)\}$, where $k_{\max}(e)$ linearly ramps from 0 to 7 over the first 50 epochs.

\subsection{Single-Pass Predictive Uncertainty for Safety Monitoring}
In safety-critical robotic interventions, assessing the \emph{trustworthiness} of a force prediction is as crucial as the prediction itself. We augment our architecture with a parallel \emph{uncertainty head} that estimates the per-axis heteroscedastic log-variance from the shared CLS token \cite{kendall2017uncertainties}:
\begin{equation}
\hat{\boldsymbol{\mu}},\;\log\hat{\boldsymbol{\sigma}}^2 = g_\theta(\mathbf{z}_{\mathrm{CLS}}),
\end{equation}
where $\hat{\boldsymbol{\mu}}\in\mathbb{R}^3$ is the force estimate and $\hat{\boldsymbol{\sigma}}^2\in\mathbb{R}^3$ is the predicted variance (ensured positive via exponentiation at runtime). During the balanced fine-tuning, we optimize the Gaussian Negative Log-Likelihood (NLL):
\begin{equation}
\mathcal{L}_{\mathrm{nll}} = \frac{1}{2}\sum_{j=1}^{3}\left(\log\hat{\sigma}_j^2 + \frac{(y_j - \hat{\mu}_j)^2}{\hat{\sigma}_j^2}\right).
\end{equation}
The final objective combines the clean and corrupted views:
\begin{equation}
\begin{aligned}
\mathcal{L}_{\mathrm{total}} = {} & \alpha\bigl[\mathcal{L}_{\mathrm{reg}}^{\mathrm{clean}} + \beta\,\mathcal{L}_{\mathrm{nll}}^{\mathrm{clean}}\bigr] \\
& + (1{-}\alpha)\bigl[\mathcal{L}_{\mathrm{reg}}^{\mathrm{corrupt}} + \beta\,\mathcal{L}_{\mathrm{nll}}^{\mathrm{corrupt}}\bigr] + \lambda\,\mathcal{L}_{\mathrm{rec}}.
\end{aligned}
\end{equation}
Evaluating this objective on both clean and corrupted views is what enables robust uncertainty calibration. The clean view anchors the predicted variance ($\hat{\sigma}^2$) at low values when the sensor is fully functional, while the corrupted view intrinsically produces larger residuals $(y_j - \hat{\mu}_j)^2$, driving the NLL loss to penalize overconfidence and output a high variance. Consequently, the network provides dynamic, physics-aware uncertainty quantification in a \emph{single forward pass}, eliminating the computational bottleneck of Deep Ensembles or MC Dropout.   

From a theoretical standpoint, the NLL objective is traditionally utilized to capture \textit{aleatoric} uncertainty, such as inherent sensor noise. However, structural sensor failures typically manifest as \textit{epistemic} uncertainty due to missing information. By explicitly enforcing the dynamic corruption curriculum during training, our dual-view objective forces the network to map severe information deficits to large residuals $(y_j - \hat{\mu}_j)^2$. As a result, the NLL head learns to output high variance for unobserved states, transforming the aleatoric formulation into an empirical proxy for epistemic structural degradation.

\begin{figure}[t]
  \centering
  \includegraphics[width=\linewidth]{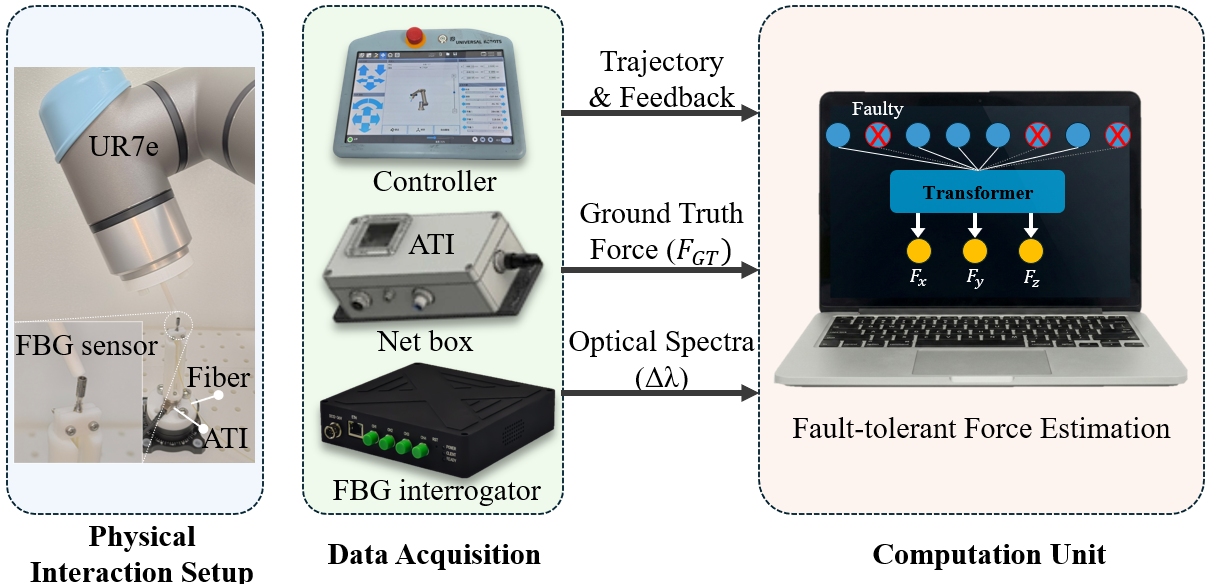}
  \caption{Hardware architecture and data flow of the experimental platform. The system integrates the physical interaction setup, synchronous data acquisition, and the real-time computation unit.}
  \label{fig:setup}
  \vspace{-8mm}
\end{figure}

\section{Experimental Setup}
\subsection{Sensor and Channel Layout}
The customized sensor prototype features eight FBG channels arranged as four fibers with two gratings per fiber, mounted around a tubular substrate to fit catheter-scale form factors.
This structure provides measurement redundancy but also introduces fiber-level failure modes.

\subsection{Calibration Platform and Data}
Figure~\ref{fig:setup} shows the calibration platform.
We use an ATI Nano17 force sensor as ground truth, a UR7e robot arm to apply controlled loads, and an FBG interrogator (100~Hz, 1~pm resolution) to acquire 8-channel signals (4 fibers $\times$ 2 gratings).
The dataset contains 35,520 training samples and 13,251 test samples.
Ground-truth forces range from approximately $[-1.0, 1.0]$~N in $F_x$, $[-1.0, 1.0]$~N in $F_y$, and $[0.0, 1.0]$~N in $F_z$ on the training set.
We further split 15\% of the training set for validation and select the best model by validation Root Mean Square Error (RMSE).
The training and test sets are collected from temporally separated loading sequences to avoid data leakage; no test-set samples appear in the training window.

\subsection{Baselines and Metrics}
We compare our method against several representative baselines from the FBG force sensing literature. To handle missing channels during inference, all single-model baselines are trained exclusively on clean data and rely on training-set mean imputation at test time. The evaluated methods include: 
(i) a classical linear least-squares calibration; 
(ii) a standard single neural network, specifically an Extreme Learning Machine (ELM);
(iii) recent meta-heuristic optimized networks, namely SHO-BPNN~\cite{li2024wavelength} and DBO-ELM~\cite{li2024modular}, where Spotted Hyena and Dung Beetle Optimizers tune the respective network hyperparameters. Although originally designed for single-channel loss scenarios ($k{=}1$), we evaluate them across all $k$; 
(iv) the Optimized-BPNN, which functions as a combinatorial model-bank following the per-pattern approach~\cite{li2023fault}. This baseline constructs an exhaustive set of 255 independent standard BPNNs, training a dedicated network for each possible non-empty channel subset using only the available valid channels as input; 
and (v) a Transformer (our architecture trained without the masked modeling curriculum) to isolate the benefits of our training strategy.

For performance evaluation, we report the test RMSE averaged over the three force axes.

\begin{figure}[t]
  \centering
  \includegraphics[width=\linewidth]{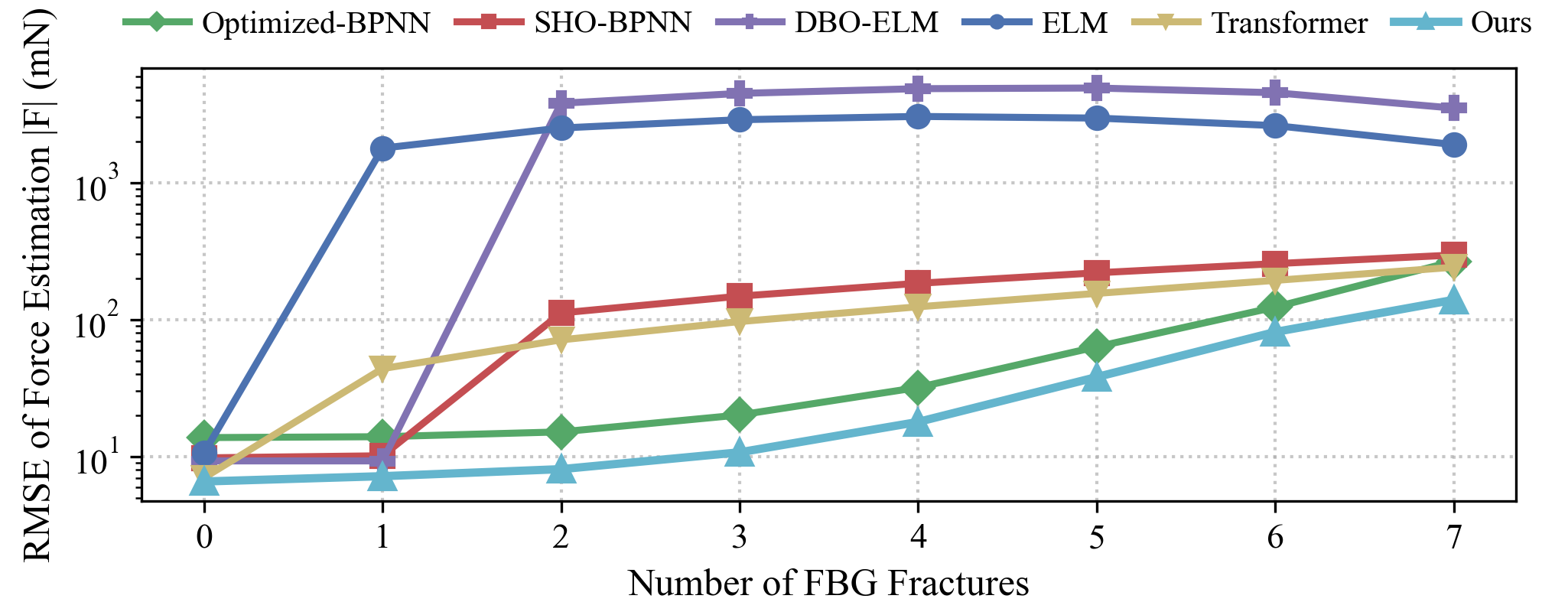}
  \caption{Comparative evaluation of test RMSE under exhaustive $k$-channel failure permutations. Note the logarithmic scale on the y-axis (mN). The proposed unified framework consistently outperforms combinatorial model-bank approaches and clean-trained neural baselines across all degradation severities.}
  \label{fig:fault_sweep}
  \vspace{-8mm}
\end{figure}

\subsection{Implementation Details}
Unless otherwise stated, the Transformer uses $d{=}512$, 8 layers, 8 heads, and SwiGLU feed-forward blocks.
We pretrain for 100 epochs with a fixed pretraining mask count of 2 channels and fine-tune for 200 epochs using AdamW ($1 \times 10^{-4}$) and batch size 1024.
Inputs are standardized using the training-set mean and standard deviation; missing channels are imputed with the training mean and indicated by the mask $\mathbf{m}$.
We set weight decay to 0, use a cosine schedule for pretraining, and a ReduceLROnPlateau schedule for fine-tuning.
The balanced fine-tuning uses equal weights for clean and corrupted views and ramps the corruption strength to $k_{\max}{=}7$.
For reproducibility, we fix all random seeds and report the exact fault protocol; code and trained checkpoints will be released upon publication.
We further evaluated training stability across five independent random seeds. Under fully functional conditions ($k=0$), the model achieves an RMSE of $0.0066 \pm 0.0001$~N. Even with four failed channels ($k=4$), the error remains as low as $0.0126 \pm 0.0004$~N, confirming the stability and low seed variance of the proposed training scheme.

\subsection{Fault Protocol}
To simulate channel failures, we exhaustively evaluate all $\binom{8}{k}$ channel masks for each $k$ ($k$-sweep) and report the average RMSE.
We also test fiber-level failures by masking each fiber group (channel pairs 1--2, 3--4, 5--6, 7--8).

\section{Results}
\subsection{Nominal and Fault-Tolerant Performance}
Figure~\ref{fig:fault_sweep} visualizes the test RMSE trends under exhaustive $k$-channel masking for $k{=}0$ through $k{=}7$.
All single-model baselines are trained on clean data only; missing channels are mean-imputed at test time.
DBO-ELM collapses once two or more channels are missing ($k \ge 2$), with RMSE exceeding 1.7~N.
Neural baselines degrade severely under four-channel failures ($k = 4$), yielding RMSE above 0.12~N.

Our balanced fine-tuning achieves the best RMSE from $k{=}1$ onward with a single unified model, while handling diverse and dynamic $k$ and providing calibrated uncertainty.
For reference, the exhaustive per-pattern BPNN bank (255 models) achieves an RMSE of 0.0114~N at $k = 0$ and 0.0154~N at $k = 4$. Under the same $k = 4$ condition, our single unified model achieves an RMSE of 0.0126~N (an 18.2\% reduction), while eliminating per-pattern calibration.
Table~\ref{tab:per_axis_k0} further breaks down the nominal ($k{=}0$) accuracy per axis, showing that our method achieves the lowest mean RMSE across all three force axes.

\begin{table}[t]
  \caption{Per-axis RMSE at $k{=}0$ (no missing channels).}
  \label{tab:per_axis_k0}
  \centering
  \scriptsize
  \setlength{\tabcolsep}{3.2pt}
  \begin{tabular}{l|cccc}
\hline
Method & RMSE$_x$ & RMSE$_y$ & RMSE$_z$ & $\overline{\mathrm{RMSE}}$ \\
\hline
Linear & 0.0508 & 0.0352 & 0.0721 & 0.0527 \\
ELM & 0.0092 & 0.0072 & 0.0103 & 0.0089 \\
Optimized-BPNN  & 0.0119 & 0.0092 & 0.0131 & 0.0114 \\
SHO-BPNN & 0.0093 & 0.0071 & 0.0088 & 0.0084 \\
DBO-ELM & 0.0078 & 0.0063 & 0.0085 & 0.0075 \\
\hline
Transformer  & 0.0071 & 0.0061 & 0.0078 & 0.0070 \\
Ours & \textbf{0.0068} & \textbf{0.0058} & \textbf{0.0071} & \textbf{0.0066} \\
\hline
\end{tabular}

  \vspace{-6mm}
\end{table}

\begin{figure}[t]
  \centering
  \includegraphics[width=\linewidth]{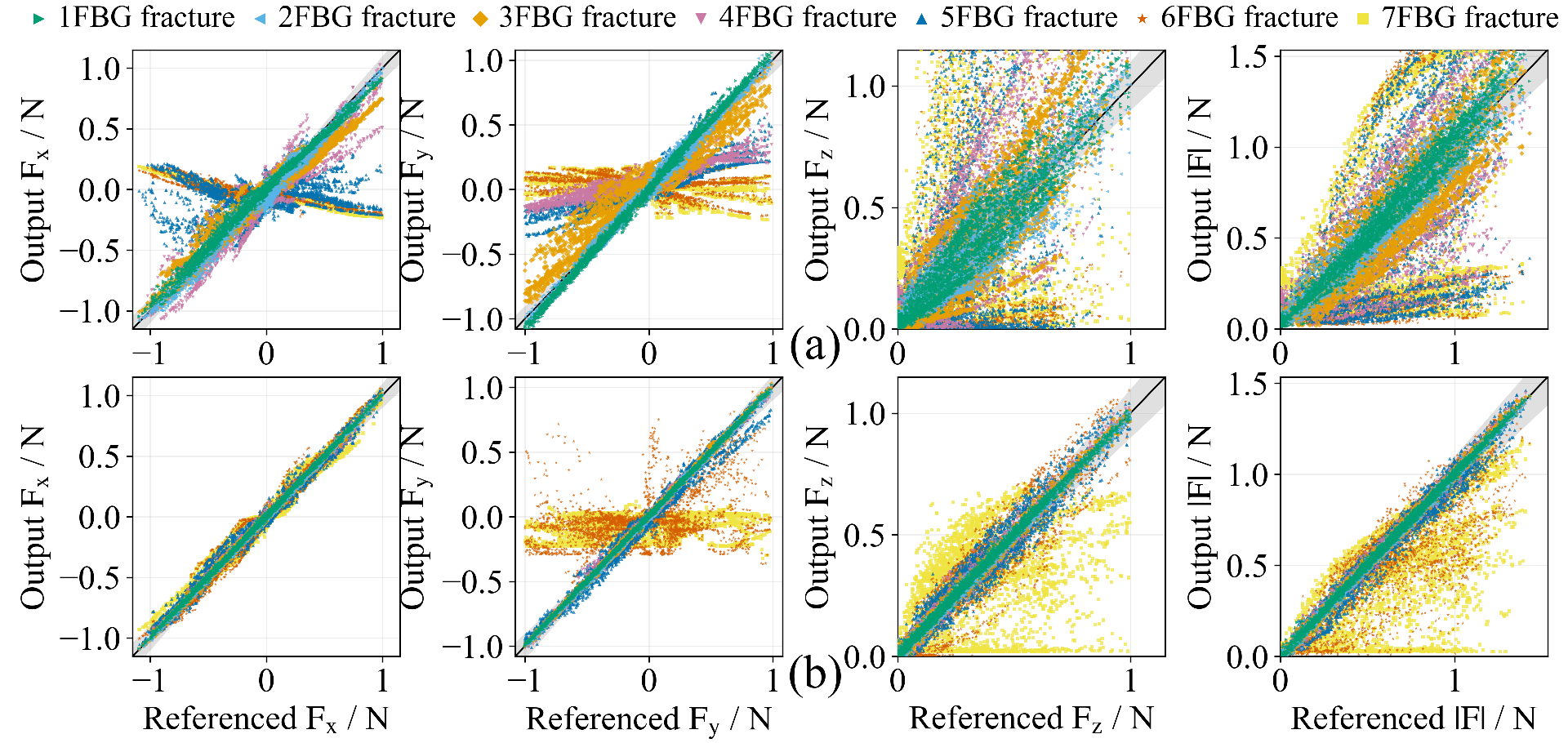}
 \caption{Predicted versus referenced forces across varying degrees of structural channel failures ($k \in [1, 7]$). (a) The Transformer baseline exhibits catastrophic estimation collapse as failures increase. (b) The proposed unified model maintains tight correlation along the ideal identity line (shaded area) even under severe physical degradation.}
  \label{fig:fig15_force}
  \vspace{-2mm}
\end{figure}

As visualized in Fig.~\ref{fig:fig15_force}(a), baseline models without fault-aware training exhibit severe predictive dispersion when subjected to progressive channel failures. Conversely, Fig.~\ref{fig:fig15_force}(b) demonstrates that the proposed fault-tolerant model maintains tight clustering around the identity line, ensuring graceful degradation and reliable force tracking even when only a fraction of the sensors remain active.

\subsection{Sensitivity and Robustness Analysis}
\begin{figure}[t]
  \centering
  \includegraphics[width=\linewidth]{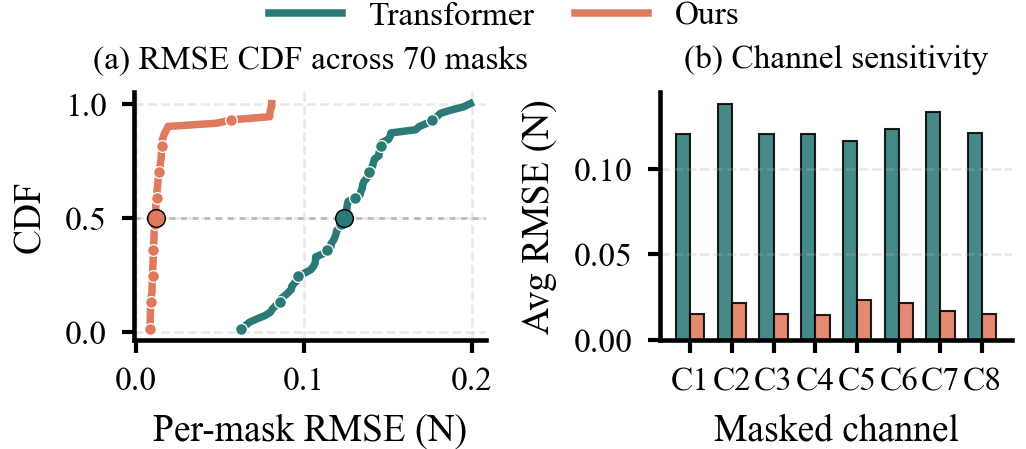}
  \caption{Exhaustive $k{=}4$ analysis: (a) CDF of per-mask RMSE over all $\binom{8}{4}$ masks; (b) per-channel sensitivity score, defined as the average RMSE across all masks containing the corresponding channel.}
  \label{fig:k4_exhaustive_summary}
  \vspace{-6mm}
\end{figure}

We next examine the worst-case regime through all $\binom{8}{4}$ masks at $k{=}4$.
Figure~\ref{fig:k4_exhaustive_summary} visualizes the per-mask error distribution Cumulative Distribution Function (CDF) and a per-channel sensitivity score computed as the average RMSE over masks that include a given channel.

We further evaluate robustness against non-missing corruptions (in normalized units): additive Gaussian noise ($\sigma{=}0.1$) and common-mode bias ($b{=}0.1$). Under severe degradation ($k{=}4$ fractures), our model maintains RMSEs of 0.0358~N (noise) and 0.0313~N (bias), compared with 0.1393~N and 0.1342~N for the Transformer.

\begin{figure}[t]
  \centering
  \includegraphics[width=\linewidth]{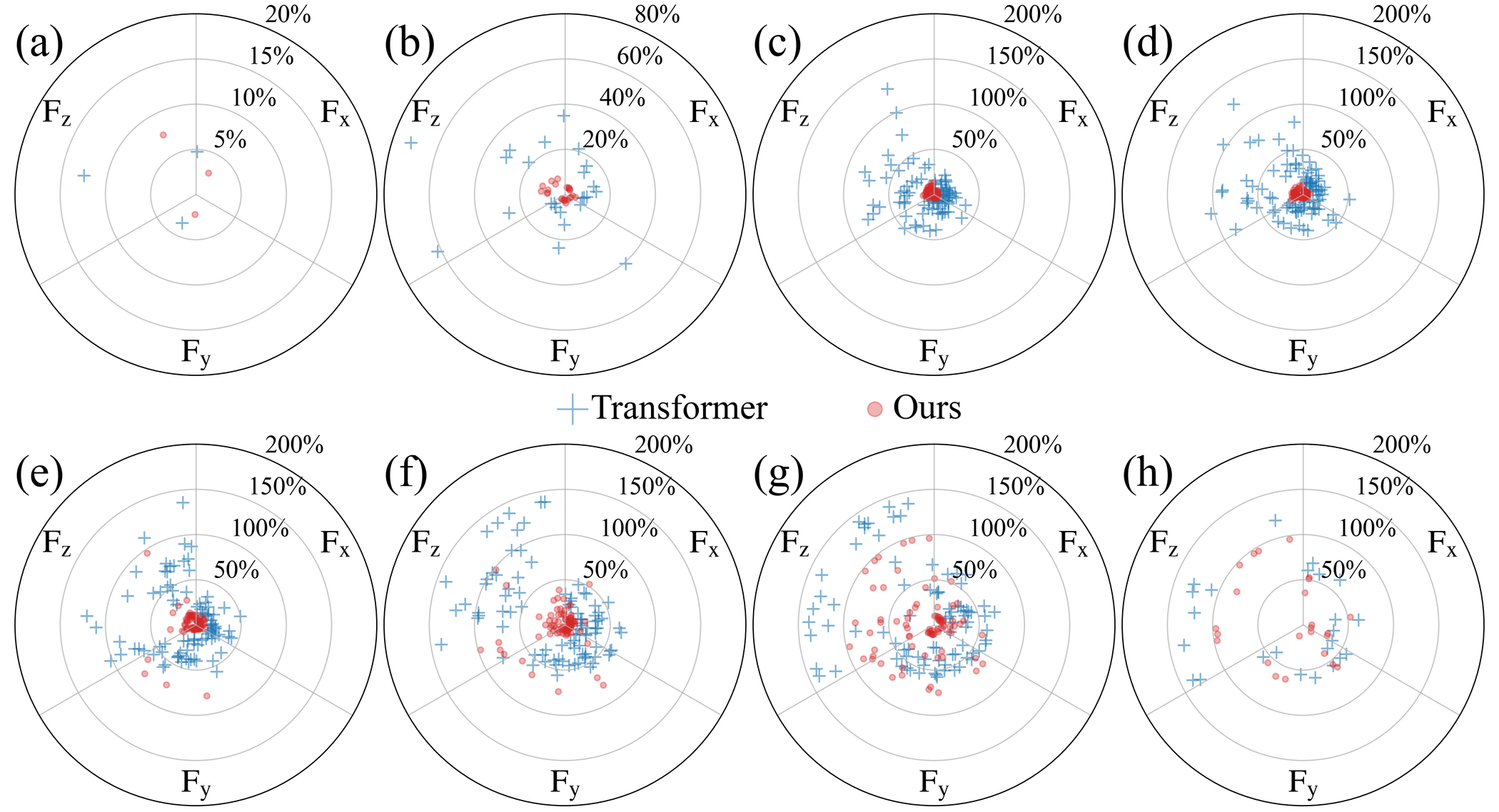}
  \caption{Type-I errors (per-axis relative deviation) under progressive channel fractures ($k=0$ to $7$). The proposed method (red dots) bounds the estimation error within tighter concentric radii than the Transformer baseline (blue crosses) across all failure severities.}
  \label{fig:type_errors}
  \vspace{-4mm}
\end{figure}

\begin{figure}[t]
  \centering
  \includegraphics[width=\linewidth]{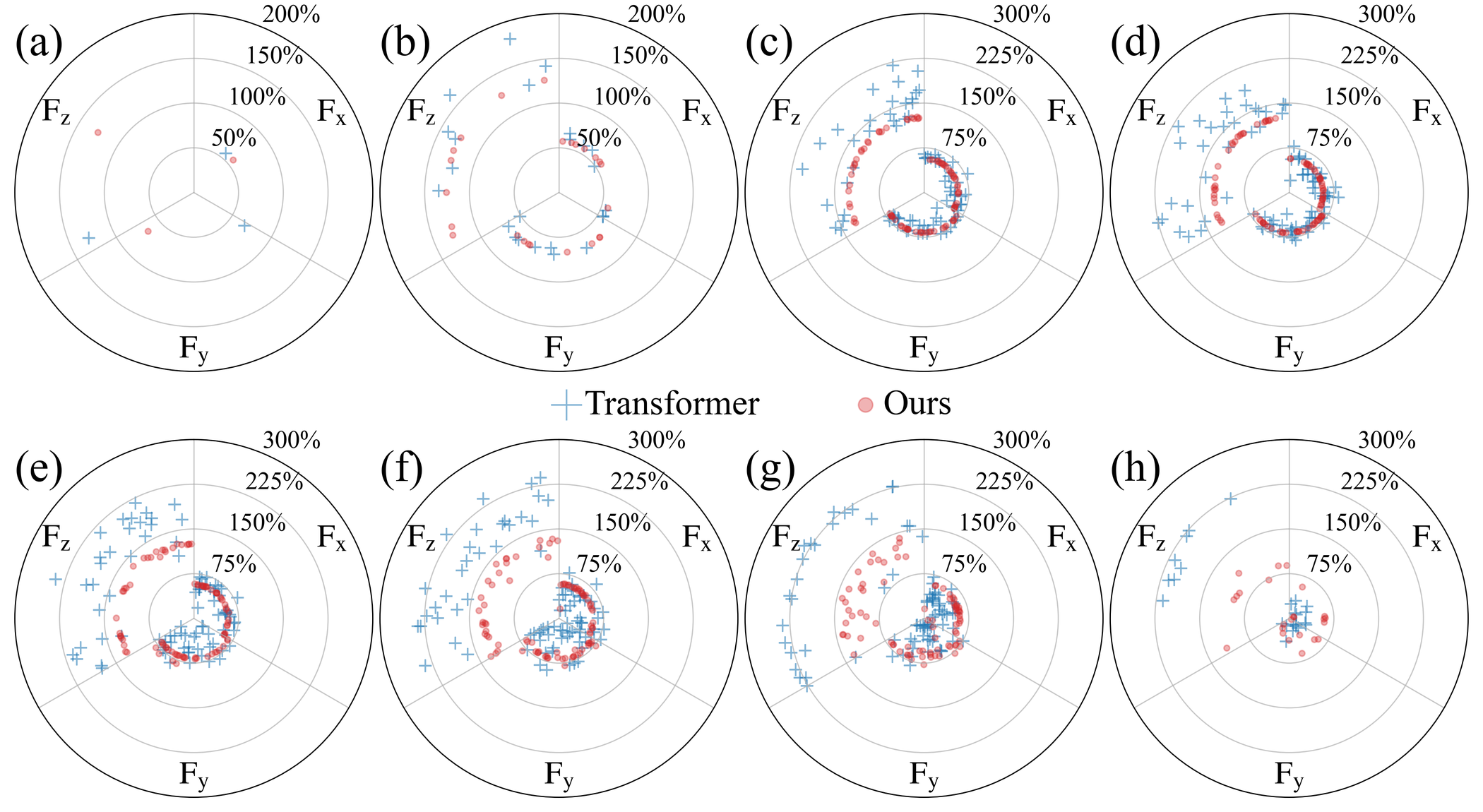}
  \caption{Type-II errors (per-axis relative deviation) under progressive channel fractures ($k=0$ to $7$).}
  \label{fig:type2_errors}
  \vspace{-8mm}
\end{figure}

Following standard force sensor calibration practice, we report the Type-I error as the per-axis relative deviation $e_{\mathrm{I},j} = (\hat{y}_j - y_j)/R_j \times 100\%$, where $R_j$ is the test-set range of axis $j$.
The Type-II (crosstalk) error uses the same metric but evaluates only the non-dominant axes: for each sample we identify the dominant loaded axis $j^* = \arg\max_j |y_j|$ and report $e_{\mathrm{II},j}$ for $j \neq j^*$.
Figure~\ref{fig:type_errors} shows that the fault-tolerant model exhibits consistently smaller Type-I error radii compared to the Transformer, with the advantage growing as more channels are lost.
Figure~\ref{fig:type2_errors} further shows the Type-II error, confirming that our model also reduces off-axis coupling under channel loss.

\subsection{Ablation}
\textbf{Component ablation.}
Table~\ref{tab:ablation} isolates the effect of key components.
Removing corruption augmentation during fine-tuning dramatically hurts robustness (``w/o corruption''), while pretraining and multi-scale tokens provide additional gains under missing channels.

\begin{table}[t]
  \caption{Ablation results (RMSE).}
  \label{tab:ablation}
  \centering
  \scriptsize
  \setlength{\tabcolsep}{3.2pt}
  \begin{tabular}{l|ccccc}
\hline
Variant & $k{=}0$ & $k{=}2$ & $k{=}4$ & $k{=}6$ & Fiber worst \\
\hline

w/o multi-scale tokens & 0.0065 & 0.0087 & 0.0192 & 0.0881 & 0.0104 \\
w/o pretrain & 0.0069 & 0.0098 & 0.0212 & 0.0910 & 0.0105 \\
w/o aux recon & 0.0066 & 0.0097 & 0.0191 & 0.0883 & 0.0103 \\
w/o corruption & 0.0073 & 0.0716 & 0.1318 & 0.1974 & 0.1170 \\
Ours & \textbf{0.0066} & \textbf{0.0086} & \textbf{0.0126} & \textbf{0.0877} & \textbf{0.0100} \\
\hline
\end{tabular}

  \vspace{-4mm}
\end{table}

We also explored a variant that explicitly masks entire fibers (two channels simultaneously) during fine-tuning. However, this yielded slightly worse nominal accuracy and $k{=}4$ performance compared to independent channel masking, leading us to adopt the latter as the default.

Beyond network architecture, we also explored the optimal corruption strength during the self-supervised pretraining phase. Empirical sweeps across masking 1 to 7 channels indicate that a moderate mask count achieves the best downstream robust RMSE. While mild masking prevents the model from relying on trivial identity mappings, excessively aggressive masking during pretraining destroys the intrinsic physical correlations among fiber strains, yielding suboptimal initialization for the regression task.

\subsection{Uncertainty Calibration and Risk-Aware Safety Contract}
We evaluate whether the predicted uncertainty $\hat{\sigma}$ provides a reliable signal for safety monitoring.

\begin{figure}[t]
  \centering
  \includegraphics[width=0.9\linewidth]{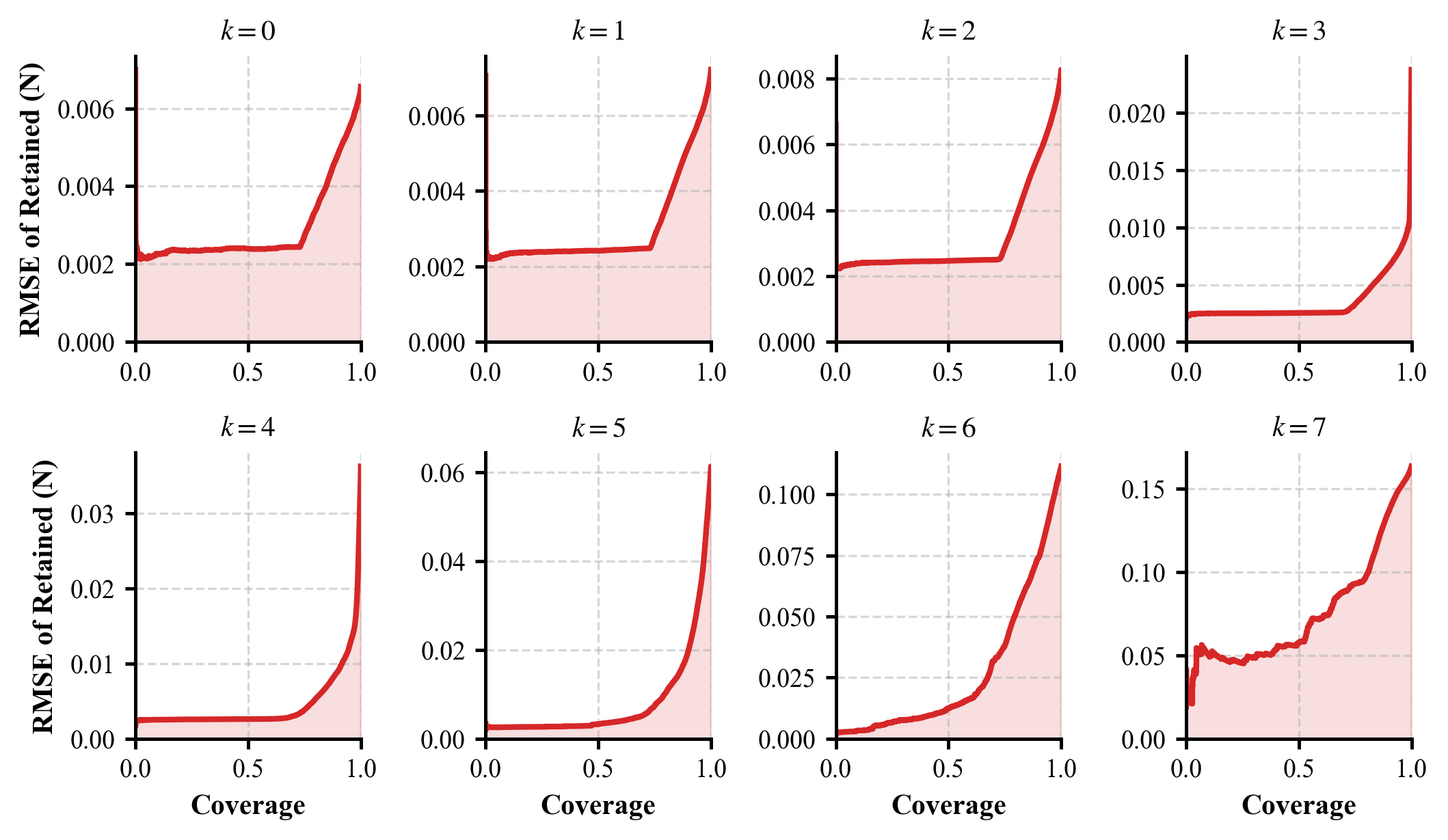}
  \caption{Risk-coverage curves: RMSE of the retained subset vs.\ coverage, for $k\in\{0,\dots,7\}$ masked channels.}
  \label{fig:risk_coverage}
  \vspace{-8mm}
\end{figure}

\begin{figure}[t]
  \centering
  \includegraphics[width=\linewidth]{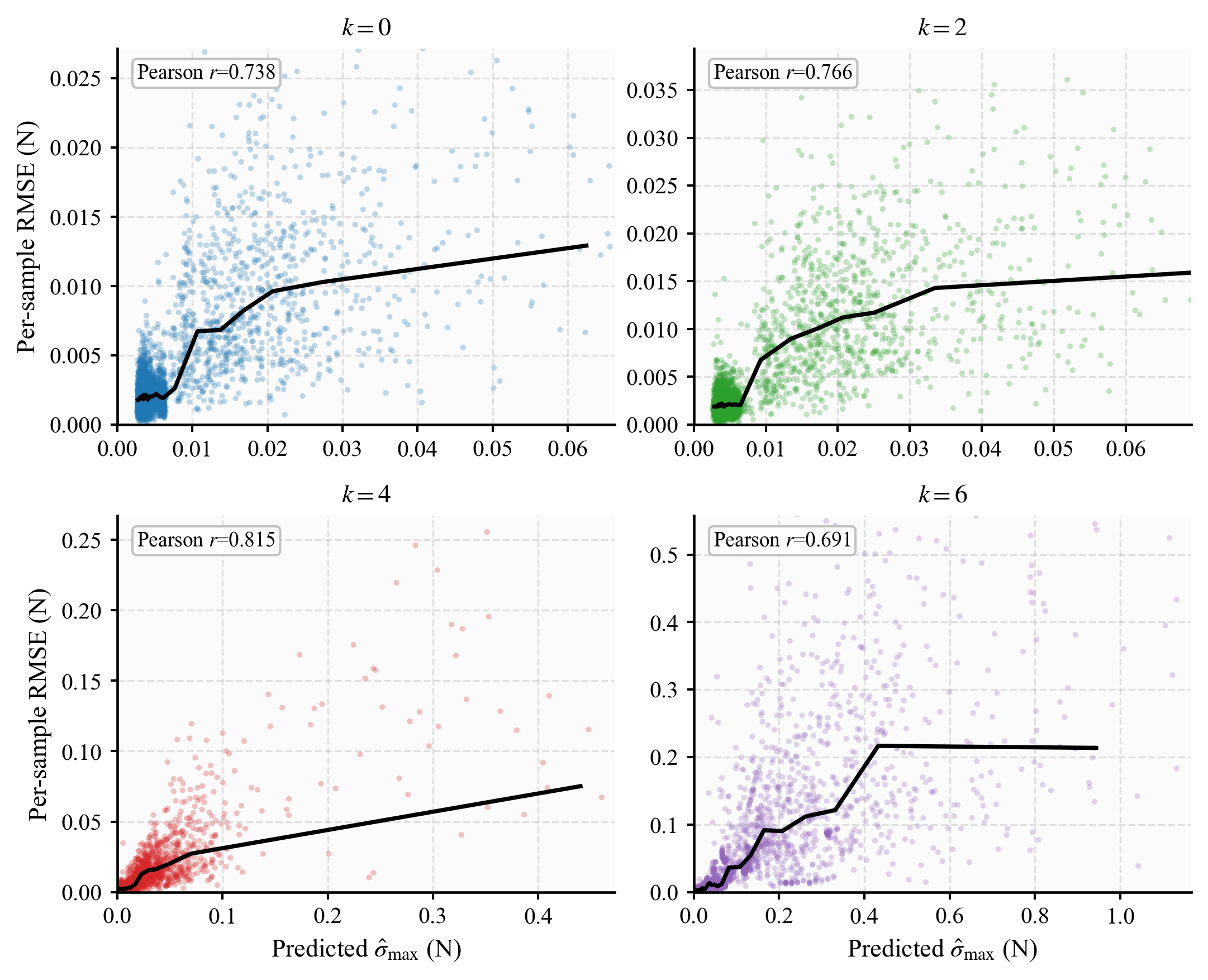}
  \caption{Correlation between the single-pass predicted uncertainty $\hat{\sigma}_{\max}$ and the actual per-sample regression RMSE under representative failure states ($k=0, 2, 4, 6$). The Pearson coefficients ($r \ge 0.69$) validate that the uncertainty head captures and scales with the actual prediction errors.}
  \label{fig:sigma_vs_error}
  \vspace{-4mm}
\end{figure}

Figure~\ref{fig:risk_coverage} shows risk-coverage curves for $k\in\{0,\dots,7\}$: we sort test samples by predicted $\hat{\sigma}_{\max}=\max_j\hat{\sigma}_j$ in ascending order and plot the RMSE of the retained (low-uncertainty) subset as a function of coverage.
At all fault severities, retaining only low-uncertainty samples yields substantially lower RMSE (Fig.~\ref{fig:risk_coverage}). Furthermore, as illustrated in Fig.~\ref{fig:sigma_vs_error}, the predicted uncertainty $\hat{\sigma}_{\max}$ correlates strongly with the per-sample RMSE ($r{=}0.738$ at $k{=}0$ up to $r{=}0.815$ at $k{=}4$), validating $\hat{\sigma}_{\max}$ as a reliable trustworthiness indicator.

We compare our NLL head against two established uncertainty quantification methods: MC Dropout~\cite{gal2016dropout} ($T{=}30$ forward passes, dropout rate $p{=}0.1$) and a Deep Ensemble~\cite{lakshminarayanan2017simple} comprising five independently trained models. To ensure a fair comparison, the MC Dropout baseline utilizes a dedicated model trained with $p{=}0.1$, while the ensemble consists of models initialized with different random seeds.

Table~\ref{tab:mc_comparison} shows that our single-pass NLL head outperforms both multi-pass baselines under nominal and degraded conditions. Under severe structural degradation ($k{=}4$), it achieves the highest Spearman correlation ($\rho{=}0.68$) and the lowest Risk-Coverage Area Under Curve (RC-AUC of 0.0046). Moreover, it delivers this uncertainty estimation at a fraction of the computational cost---$1/5$ that of the Deep Ensemble and $1/30$ that of MC Dropout---making it well suited to real-time, high-frequency robotic control loops.

\begin{table}[t]
  \caption{Uncertainty method comparison.}
  \label{tab:mc_comparison}
  \centering
  \scriptsize
  \setlength{\tabcolsep}{3.2pt}
  \begin{tabular}{l|cc|cc|c}
  \hline
  & \multicolumn{2}{c|}{$k{=}0$} & \multicolumn{2}{c|}{$k{=}4$} & Inference \\
  Method & $\rho$ & RC-AUC & $\rho$ & RC-AUC & cost \\
  \hline
  MC Dropout ($T{=}30$) & 0.57 & 0.0032 & 0.63 & 0.0061 & $30\times$ \\
  Deep Ensemble ($5\times$) & 0.56 & 0.0031 & 0.63 & 0.0047 & $5\times$ \\
  NLL (ours) & \textbf{0.58} & \textbf{0.0028} & \textbf{0.68} & \textbf{0.0046} & $1\times$ \\
  \hline
  \end{tabular}
  \vspace{-6mm}
\end{table}

\begin{figure}[t]
  \centering
  \includegraphics[width=0.9\linewidth]{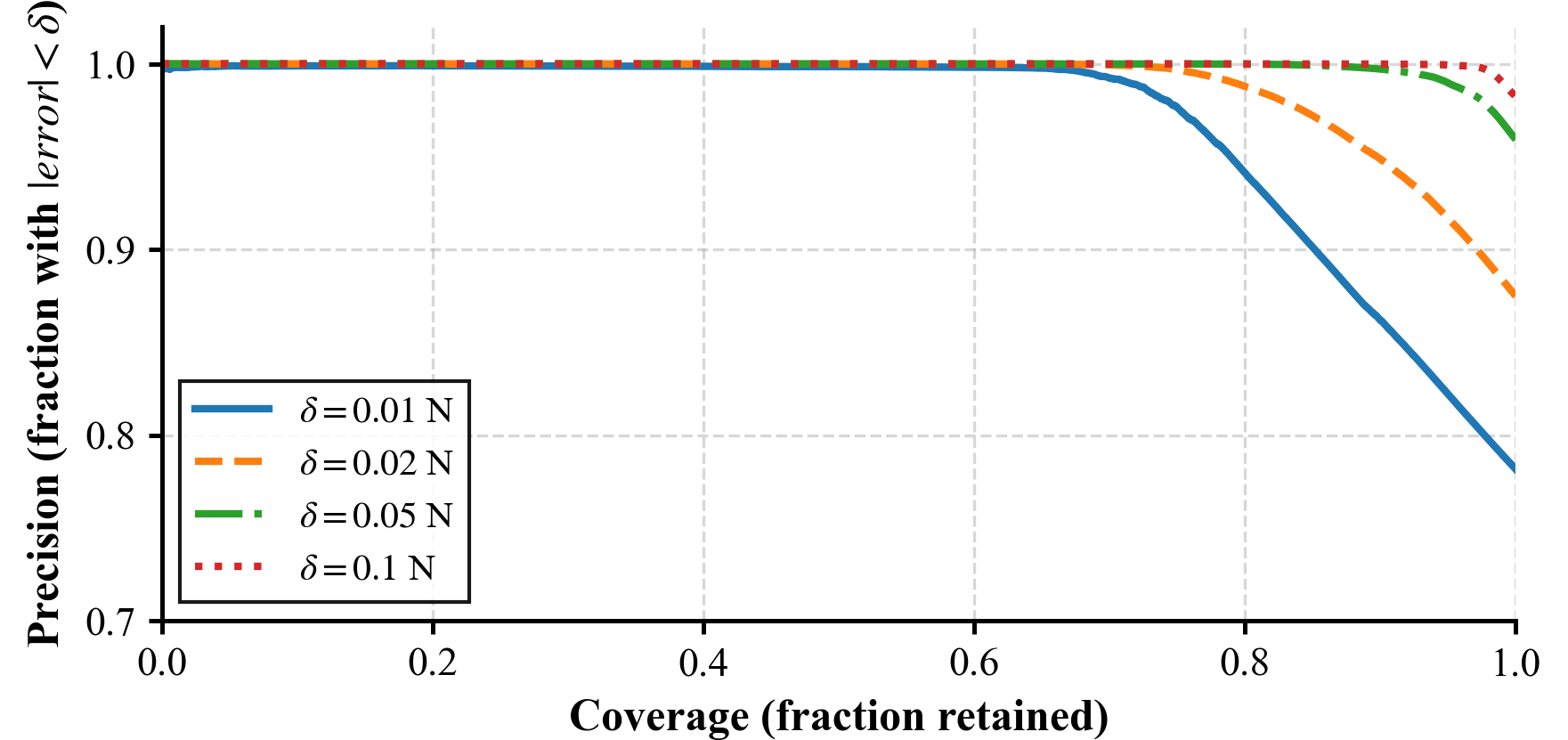}
  \caption{Safety contract: precision (fraction with $|\mathrm{error}|<\delta$) vs.\ coverage at $k{=}4$, for several error thresholds $\delta$.}
  \label{fig:contract}
  \vspace{-2mm}
\end{figure}

For deployment, we establish a safety contract: given a threshold $\tau$ on $\hat{\sigma}_{\max}$, what fraction of predictions can be trusted to have an error strictly below $\delta$~N? Figure~\ref{fig:contract} illustrates this critical trade-off between system availability (Coverage) and force estimation reliability (Precision) under the severe $k{=}4$ fracture scenario. 
When absolute clinical safety requires the error to be tightly bounded ($\delta{=}0.02$~N), it is practically infeasible to guarantee 100\% reliability under such extreme structural failure (the rightmost tail of the orange curve drops to ${\sim}88\%$). However, using the predicted uncertainty, we can trade availability for safety. For instance, setting a warning threshold ($\tau_1{=}0.022$~N) sacrifices 31\% of the system's availability (Coverage drops to 69\%) to intercept unreliable predictions; in return, the retained predictions attain a precision of 98.5\% (Nominal state). 
Similarly, relaxing the threshold to $\tau_2{=}0.044$~N expands the coverage to 82\% while still maintaining a 92.3\% precision (Degraded state). Any prediction with $\hat{\sigma}_{\max}{\geq}\tau_2$ forces the robot into an emergency Stop, preventing unsafe tissue interactions.
This procedure is analogous to selecting a classification threshold on a validation Receiver Operating Characteristic (ROC) curve; it requires per-sensor threshold recalibration, and cross-sensor robustness of the threshold selection is left for future work.

\begin{figure}[t]
  \centering
  \includegraphics[width=0.9\linewidth]{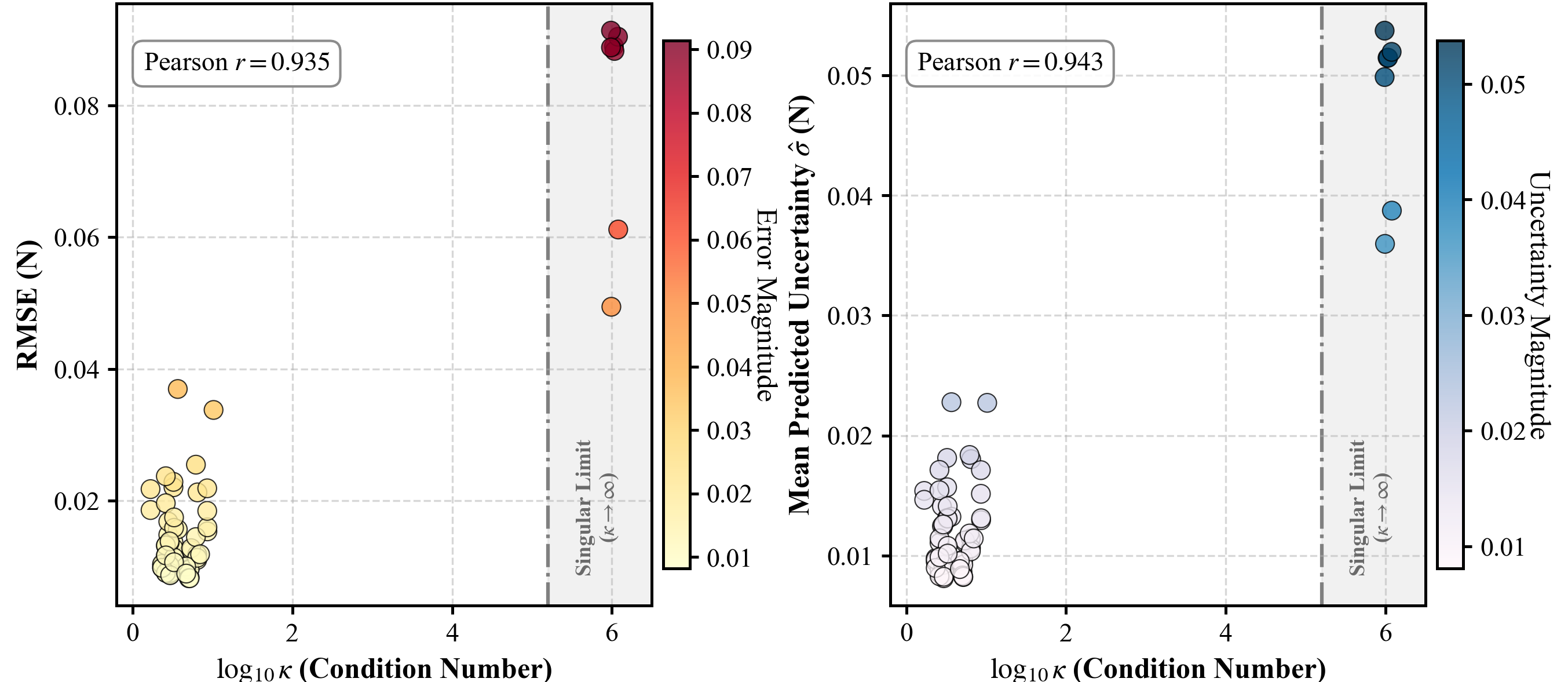}
  \caption{Condition number $\kappa$ vs.\ RMSE (left) and vs.\ predicted $\hat{\sigma}$ (right) for all 70 exhaustive $k{=}4$ masks.}
  \label{fig:kappa}
  \vspace{-8mm}
\end{figure}

\looseness=-1
To understand \emph{why} certain mask patterns are harder, we fit a linear sensitivity matrix $\mathbf{S}\in\mathbb{R}^{8\times 3}$ from training data via ordinary least-squares regression and compute the condition number $\kappa$ of the sub-matrix $\mathbf{S}_{\mathrm{sub}}$ for each of the $\binom{8}{4}$ masks at $k{=}4$.
Figure~\ref{fig:kappa} shows that $\log\kappa$ correlates strongly with RMSE (Pearson $r{=}0.935$): masks that leave a poorly conditioned (near-singular) channel subset produce higher errors.

Notably, 7 out of 70 masks yield singular sub-matrices ($\kappa\to\infty$, plotted separately). The 5 worst-performing masks (RMSE ${\sim}0.07$~N vs. a $0.01$~N median) all belong to this singular group and consistently involve the simultaneous masking of fiber 3, highlighting its unique role in structural observability. The predicted uncertainty also tracks $\log\kappa$ ($r{=}0.943$), indicating that the network has implicitly learned the physical condition-number landscape without explicit supervision.

\subsection{In-Vivo Validation Under Real Fiber Fractures}
\label{sec:invivo}

We validated the proposed method beyond simulated data masking through \textit{in vivo} experiments on a porcine model, deploying the catheter-scale FBG sensor during unstructured tissue interactions.

All animal experiments were conducted in accordance with the relevant ethical guidelines and regulations, and were approved by the Institutional Animal Care and Use Committee (IACUC) at the authors' institution.

\begin{figure}[t]
    \centering
    \includegraphics[width=\linewidth]{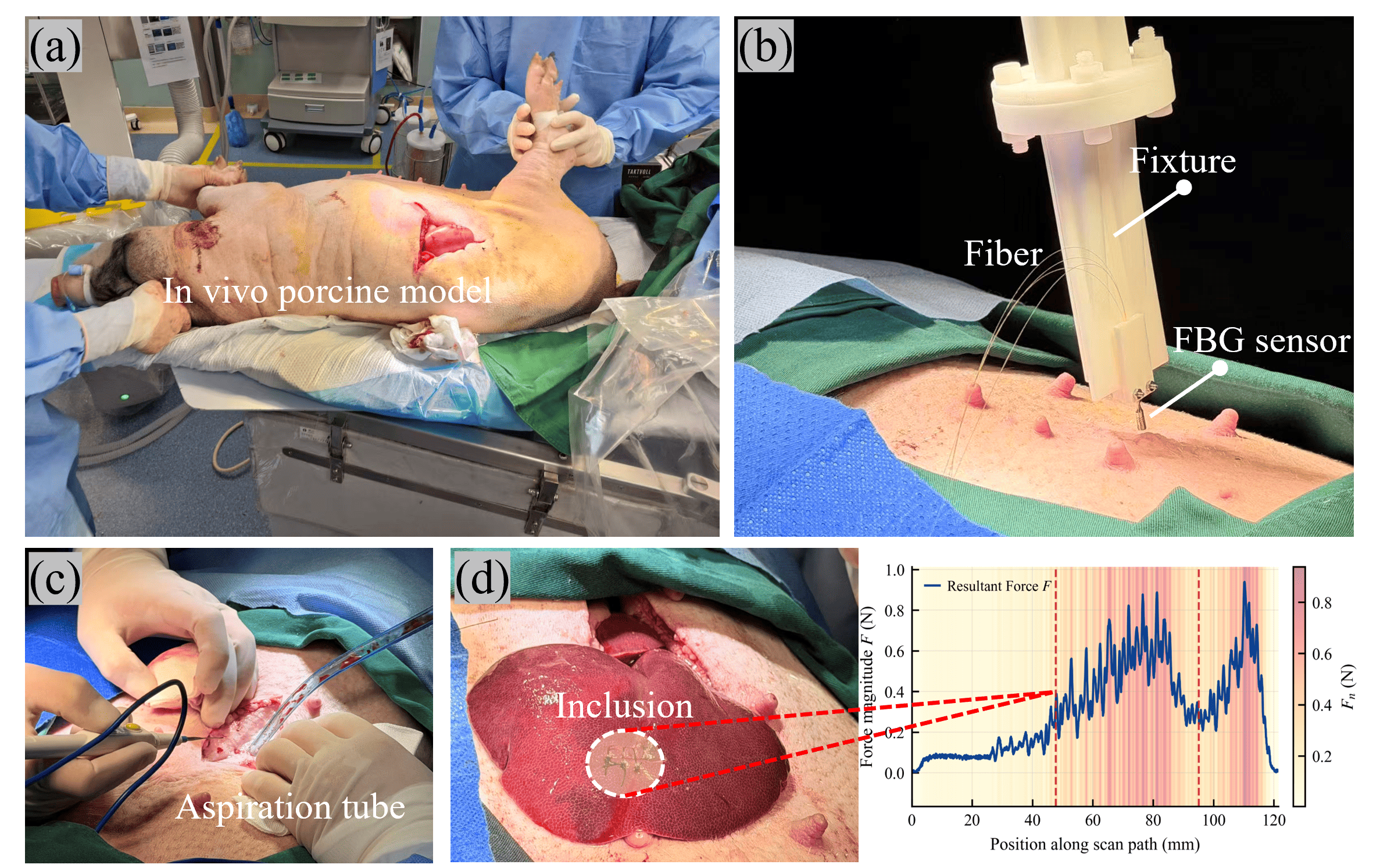}
    \caption{Experimental setup and procedure for the \textit{in vivo} porcine model study. 
    (a) Overall view of the surgical field on the living pig. 
    (b) Close-up view of the palpation platform, illustrating the fixture, optical fibers, and the FBG sensor interacting with the biological tissue. 
    (c) Surgical preparation utilizing an aspiration tube. 
    (d) Exposed liver tissue with the target inclusion area highlighted by the dashed circle. The right panel illustrates the resultant force profile as the sensor scans across the inclusion path, clearly capturing the structural stiffness of the embedded tumor.}
    \label{fig:invivo}
    \vspace{-2mm}
\end{figure}

Figure~\ref{fig:invivo} illustrates the experimental setup and the precise surgical procedure. To emulate a clinically relevant palpation task, an artificial tumor (rigid inclusion) was surgically implanted into the porcine liver parenchyma (Fig.~\ref{fig:invivo}(d)). The sensor then performed automated palpation trajectories across the target region. Figure~\ref{fig:invivo}(a--b) details the \textit{in vivo} surgical field and the interaction between the optical FBG sensor and the biological tissue. Rather than relying solely on software-level signal masking, we induced physical fiber fractures at the hardware connector during the experiment. This created unrecoverable channel-loss scenarios ($k{=}2$, $4$, and $6$). These physical experiments introduce realistic hardware-level effects---such as optoelectronic noise and baseline shifts---that are difficult to fully capture via software simulation.

\subsubsection{In-Vivo Palpation and Target Detection}
As shown in the spatial response plot adjacent to Fig.~\ref{fig:invivo}(d), the fully intact sensor ($k{=}0$) clearly captures a distinct force peak as it traverses the embedded inclusion. This behavior is consistent with the expected mechanical profile of a hidden stiff nodule, suggesting that the sensor can capture tissue stiffness variations in a realistic surgical environment.

\subsubsection{Dynamic Safety Policy Activation under Hardware Fractures}
\begin{figure}[t]
    \centering
    \includegraphics[width=\linewidth]{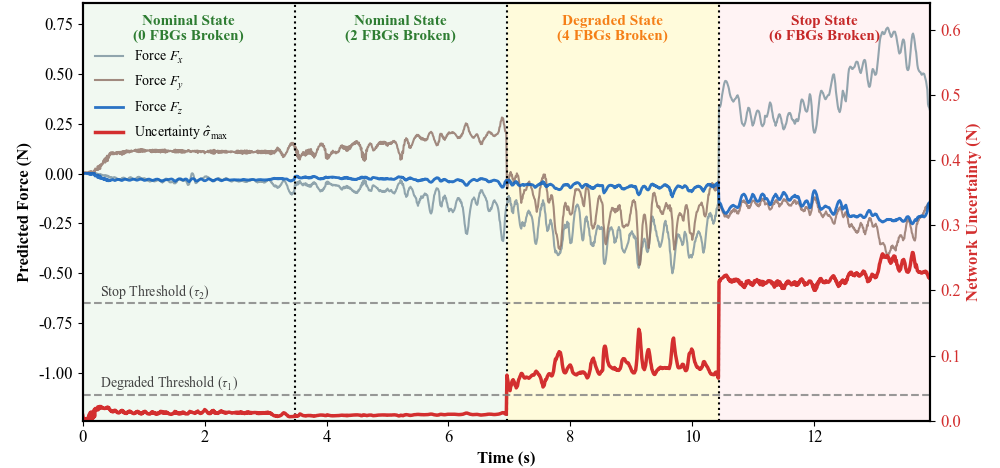} 
    \caption{Dynamic transition of robotic safety states triggered by progressive physical fiber fractures during the \textit{in vivo} experiment. The real-time predictive uncertainty ($\hat{\sigma}_{\max}$, thick red curve) reliably monitors structural health: it maintains the system in the \textit{Nominal State} for minor damages ($k \le 2$), crosses the degraded threshold ($\tau_1$) to trigger a \textit{Degraded State} warning upon severe information loss ($k=4$), and finally breaches the stop threshold ($\tau_2$) to enforce a \textit{Stop State} before catastrophic, unobservable force oscillations ($k=6$) interact with the biological tissue.}
    \label{fig:force_uncertainty_states}
    \vspace{-8mm}
\end{figure}

Beyond nominal tracking, the ultimate goal of our framework is to preserve surgical safety under structural degradation. Figure~\ref{fig:force_uncertainty_states} demonstrates the real-time temporal response of our safety policy as physical fractures progressively worsen. The force predictions ($F_x, F_y, F_z$) and the network's predictive uncertainty ($\hat{\sigma}_{\max}$) are plotted over time.

Initially, with the sensor fully intact ($k{=}0$), the uncertainty remains low, keeping the system within the \textit{Nominal State}. When a partial physical fracture occurs ($k{=}2$), the fault-tolerant model maintains stable force tracking; the uncertainty $\hat{\sigma}_{\max}$ exhibits a slight increase but remains safely below the Degraded Threshold ($\tau_1$).

However, as the structural integrity critically drops ($k{=}4$), the unobservable estimation error rises due to severe information loss, causing the force outputs to become highly oscillatory. The uncertainty head reflects this degradation in real time: $\hat{\sigma}_{\max}$ rises above $\tau_1$, transitioning the system into the \textit{Degraded State}. Finally, under a catastrophic hardware failure ($k{=}6$), the predicted uncertainty exceeds the Stop Threshold ($\tau_2$), triggering the \textit{Stop State}. This behavior demonstrates that $\hat{\sigma}_{\max}$ serves as a practical and responsive proxy for unobservable prediction errors in clinical settings.

\section{Conclusion}
We presented a unified mask-aware Transformer framework for fault-tolerant three-axis force estimation from multi-channel FBG sensors. The key contributions are threefold. First, by explicitly incorporating observation masks, the single architecture handles diverse and dynamic channel-loss patterns, eliminating the $2^C{-}1$ exponential scaling overhead of conventional model banks. Second, integrating self-supervised masked reconstruction with a balanced corruption curriculum enables the model to maintain an RMSE below 0.013~N under severe 4-channel failures, significantly outperforming existing model-bank approaches while bypassing pattern-specific calibration. Third, we proposed a single-pass heteroscedastic uncertainty head that strongly correlates with the physical ill-conditioning of the sensor structure (Pearson $r{=}0.943$). 

This uncertainty quantification translates directly into an actionable $\tau$--$\delta$ safety contract for surgical robotics. As demonstrated in our \textit{in vivo} porcine model study, the real-time three-state safety policy (Nominal/Degraded/Stop) maintains usable force tracking under minor damage ($k{\le}2$) and safely triggers a system halt prior to catastrophic structural failure ($k{=}6$). Ultimately, this framework demonstrates the risk-aware resilience necessary for developing safer, closed-loop force control systems in complex minimally invasive interventions.

\bibliographystyle{IEEEtran}
\bibliography{references}

\end{document}